\documentclass[letterpaper, 10 pt, conference]{ieeeconf}  % Comment this line out if you need a4paper

\IEEEoverridecommandlockouts                              % This command is only needed if 
\usepackage{graphicx}
\usepackage{amsmath} % assumes amsmath package installed
\usepackage{amssymb}  % assumes amsmath package installed
\usepackage{url}

\title{\LARGE \bf
Autonomous Vineyard Exploration using LiDAR Sensing:\\A Field Evaluation on Legged-Wheeled Quadruped Robot
}

\author{Spencer Sweeney$^{1}$, Hongbo Li$^{2}$, Michael Ziegltrum$^{2}$, Dimitra Koumoutsou$^{2,3}$,\\ Andromachi Maria Delfaki$^{2}$, Taskin Padir$^{1}$, Dimitrios Kanoulas$^{2,3}$
\thanks{$^{1}$Institute for Experiential Robotics, Northeastern University, Boston, Massachusetts, USA.}%
\thanks{$^{2}$Department of Computer Science, University College London, Gower Street, WC1E 6BT, London, UK. {\tt\small d.kanoulas@ucl.ac.uk}}%
\thanks{$^{3}$Archimedes/Athena RC, Greece.}%
\thanks{This work was supported by the UKRI FLF [MR/V025333/1] (RoboHike) and partially supported by project MIS 5154714 of the National Recovery and Resilience Plan Greece 2.0 funded
by the European Union under the NextGenerationEU Program.  For the purpose of Open Access, the author has applied a CC BY public copyright license to any Author Accepted Manuscript version arising from this submission.}}

\begin{document}

\maketitle
\thispagestyle{empty}
\pagestyle{empty}

%%%%%%%%%%%%%%%%%%%%%%%%%%%%%%%%%%%%%%%%%%%%%%%%%%%%%%%%%%%%%%%%%%%%%%%%%%%%%%%%
\begin{abstract}
Autonomous navigation in vineyards presents significant challenges due to uneven terrain, sparse vegetation, and the repetitive geometric nature of vineyard rows. This paper evaluates a LiDAR-based autonomous exploration system deployed on a Unitree Go2 Wheeled quadruped robot. Our approach utilizes an iterative RANSAC-based line detection pipeline to identify vineyard rows and a finite-state navigation strategy to govern row selection and traversal. Field evaluations conducted at Denbies Vineyard (UK) demonstrate the system's ability to perform sustained autonomous row following, completing a $370\,\mathrm{m}$ up-and-down traversal without intervention. We analyze the failure modes encountered, specifically focusing on line detection instability in sparse vegetation and the trade-offs of spatial filtering. The results suggest that, while RANSAC provides a robust geometric primitive, additional temporal filtering and adaptive perception are required for reliable lane-to-lane transitions in commercial vineyard environments. A video of the experimental results is available at: \url{youtu.be/1om1HFGF4ps}

\end{abstract}

%%%%%%%%%%%%%%%%%%%%%%%%%%%%%%%%%%%%%%%%%%%%%%%%%%%%%%%%%%%%%%%%%%%%%%%%%%%%%%%%

\section{INTRODUCTION}

Autonomous navigation is a key enabling capability for agricultural robots tasked with monitoring, inspection, and field operations, but reliable autonomy remains difficult in unstructured outdoor environments due to terrain variability, sensing uncertainty, and seasonal change~\cite{bechar2016agrirobots,bac2014harvesting,sebastian2025stepp}. Vineyards present a particularly demanding setting: rows are narrow and repetitive, terrain is often sloped and muddy, and vegetation structure varies strongly across growth stages and management practices.

A range of sensing modalities have been explored for row-based agricultural navigation. GNSS/RTK solutions offer high accuracy but require costly infrastructure and prior surveying~\cite{lee2024compact}. Vision-based approaches exploit appearance cues from monocular or RGB-D cameras but remain sensitive to illumination, shadows, and seasonal canopy changes~\cite{mendez2023vision}. LiDAR, by contrast, provides direct geometric measurements that are largely invariant to lighting conditions, making it suitable for the high variability of the appearance of outdoor agricultural environments~\cite{shi2023row}. LiDAR-based perception has also been shown to support real-time mapping and terrain understanding at scale in unstructured outdoor environments~\cite{jiao2025metricsemantic}, further motivating its use as the primary perception modality for field deployment.

Despite recent progress in LiDAR-based vineyard navigation, there remains limited reporting on deployment-time failure modes and perception--planning interactions during extended real-world operation, particularly under late-season conditions where row boundaries may only be intermittently observable.

In this work, we evaluate a LiDAR-based autonomous vineyard exploration system deployed on a Unitree Go2-W wheeled quadruped. Rather than proposing a new row-detection primitive, this paper contributes a deployment-driven evaluation: (i)~we demonstrate sustained autonomous in-row navigation in a commercial vineyard, including a fully autonomous $370\,\mathrm{m}$ up-and-down traversal; (ii)~we analyze failure modes arising from intermittent observability under sparse vegetation; and (iii)~we examine mitigation strategies and their trade-offs between in-row stability and lane-to-lane exploratory capability.

\begin{figure}[t]
  \centering
  \includegraphics[width=\linewidth]{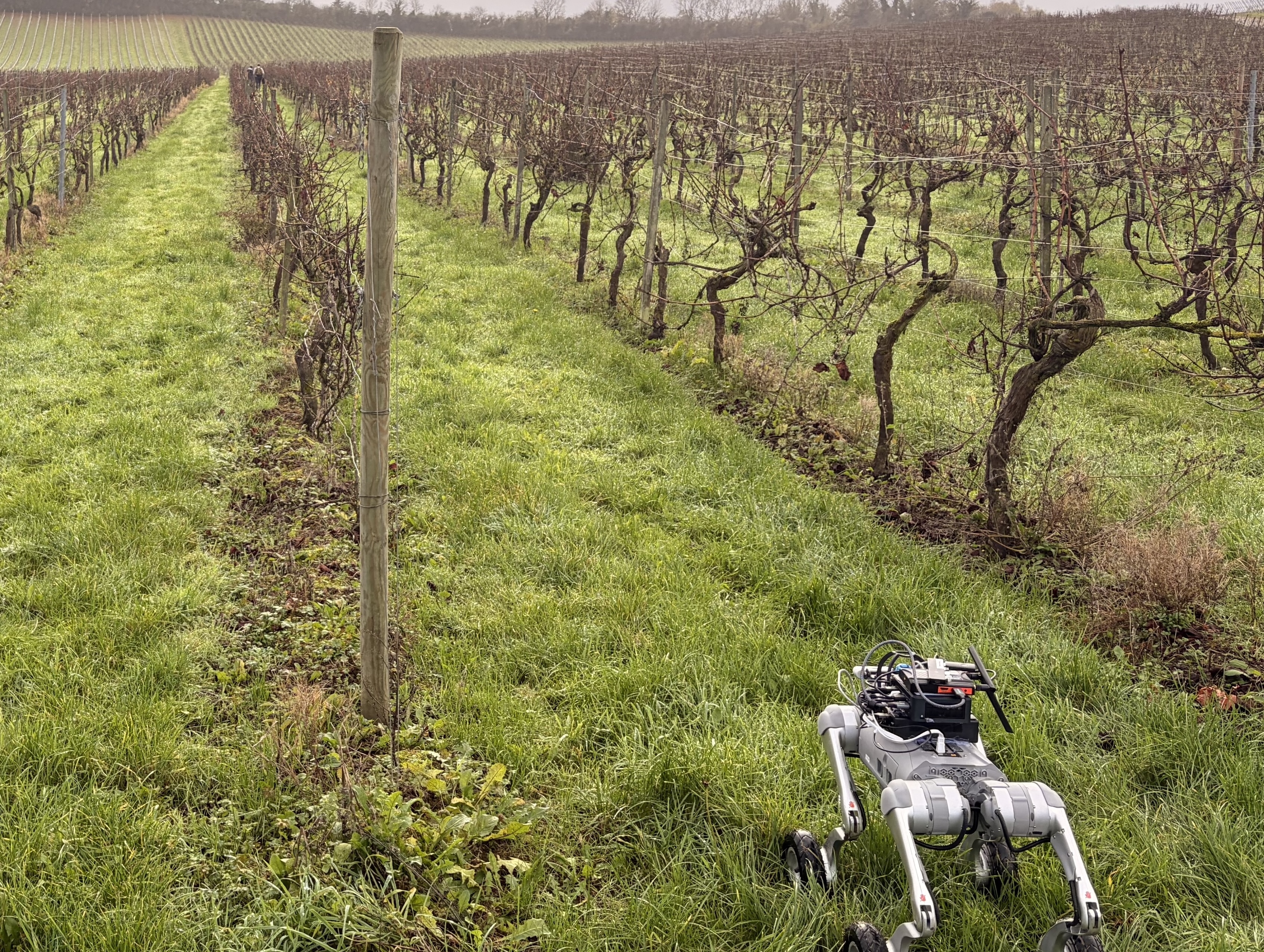}
  \caption{Mobile robot platform during autonomous testing at Denbies Vineyard (UK)
  on November 28, 2025.}
  \label{fig:platform_in_vineyard}
\end{figure}

\section{RELATED WORK}
Crop row detection is a foundational capability for agricultural robot guidance and field operations across a range of sensing modalities and crop types. Shi~\textit{et al.}~\cite{shi2023row} provide a comprehensive survey of row-detection--based navigation methods, highlighting trade-offs between vision-based, LiDAR-based, and multi-sensor approaches and the challenges posed by environmental variability and crop growth stage. LiDAR-only row navigation has also been demonstrated in broader row-crop settings, reinforcing the appeal of geometry-driven approaches when GPS is unreliable or unavailable~\cite{malavazi2018lidar}.

Within vineyard environments specifically, two works are most closely related to ours. Nehme~\textit{et al.}~\cite{nehme2021lidar} apply Hough transform--based regression to LiDAR data for in-row structure tracking, demonstrating robust vineyard navigation validated against RTK-GNSS ground truth without prior field surveying. Bertoglio~\textit{et al.} present a map-free LiDAR-only system capable of in-row traversal, turning, and end-row maneuvers using only geometric and proprioceptive cues~\cite{bertoglio2023mapfree}. Both works validate LiDAR-based navigation in real vineyard deployments, but focus primarily on navigation accuracy rather than deployment-time failure modes under adverse seasonal conditions.

\section{SYSTEM \& METHODOLOGY}
\begin{figure}[t]
  \centering
  \includegraphics[width=\linewidth]{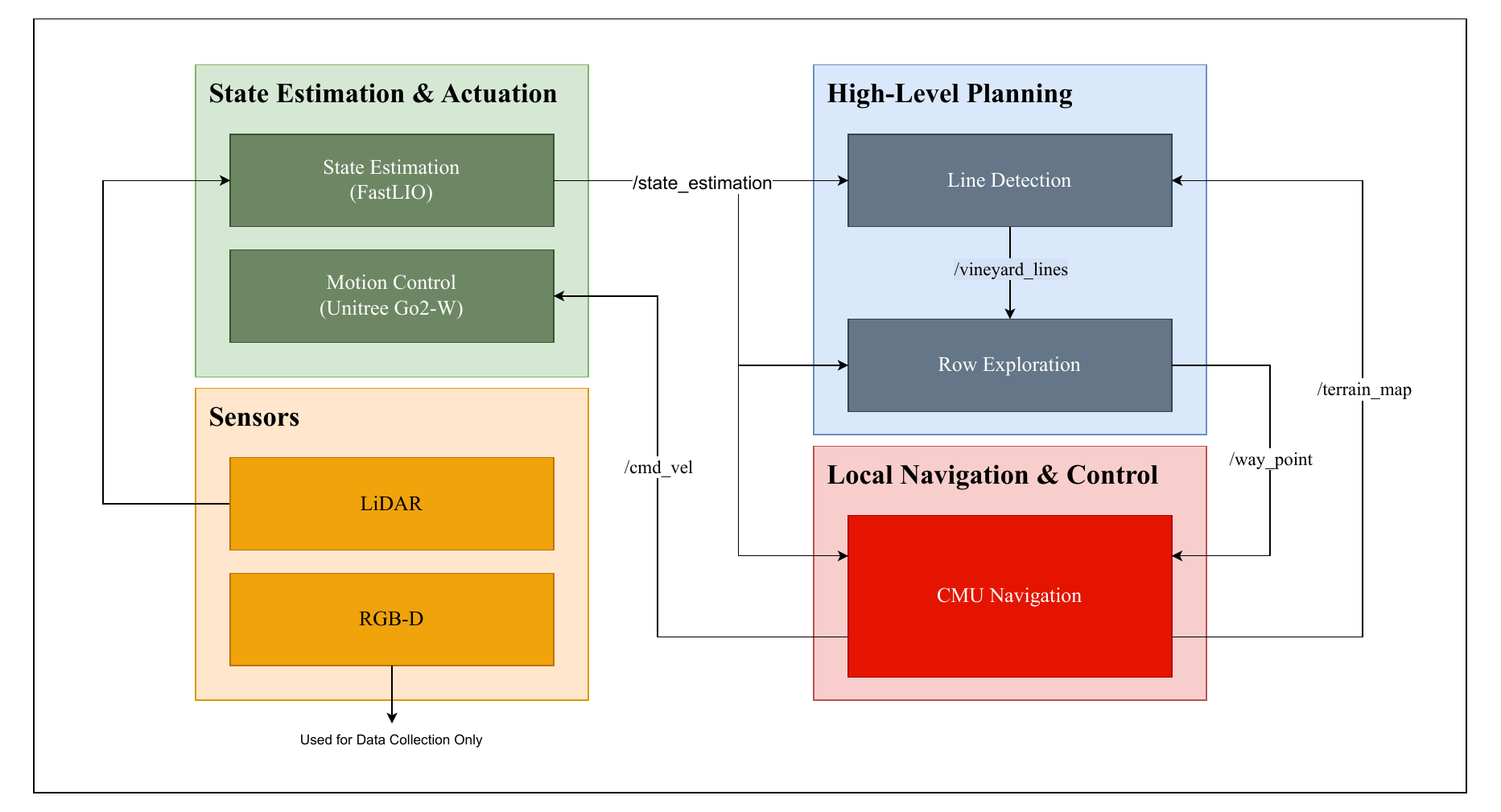}
  \caption{System architecture. LiDAR feeds FAST-LIO for state estimation and the
  Line Detection module, whose output drives the Row Exploration planner.
  Waypoints are executed by the CMU Navigation stack.}
  \label{fig:system_diagram}
\end{figure}

The system is built on a Unitree Go2-W wheeled quadruped equipped with a Livox Mid-360
LiDAR ($360^\circ{\times}59^\circ$ FoV, $200\,\mathrm{k\,pts/s}$, ${\leq}2\,\mathrm{cm}$
precision at $10\,\mathrm{m}$) as the primary perception sensor, and an Intel RealSense
D435i used for data collection only. All computation runs onboard in ROS Noetic on
Ubuntu 20.04. As shown in Fig.~\ref{fig:system_diagram}, LiDAR point clouds feed
FAST-LIO~\cite{xu2021fastlio} for real-time pose estimation, and the CMU Navigation
stack~\cite{cao2022cmu_nav} for terrain-aware local planning and collision avoidance.
High-level vineyard reasoning is handled by two custom modules: Line Detection and
Row Exploration, described below.

\subsection{LiDAR-Based Line Detection}

\begin{figure}[t]
  \centering
  \includegraphics[width=\linewidth]{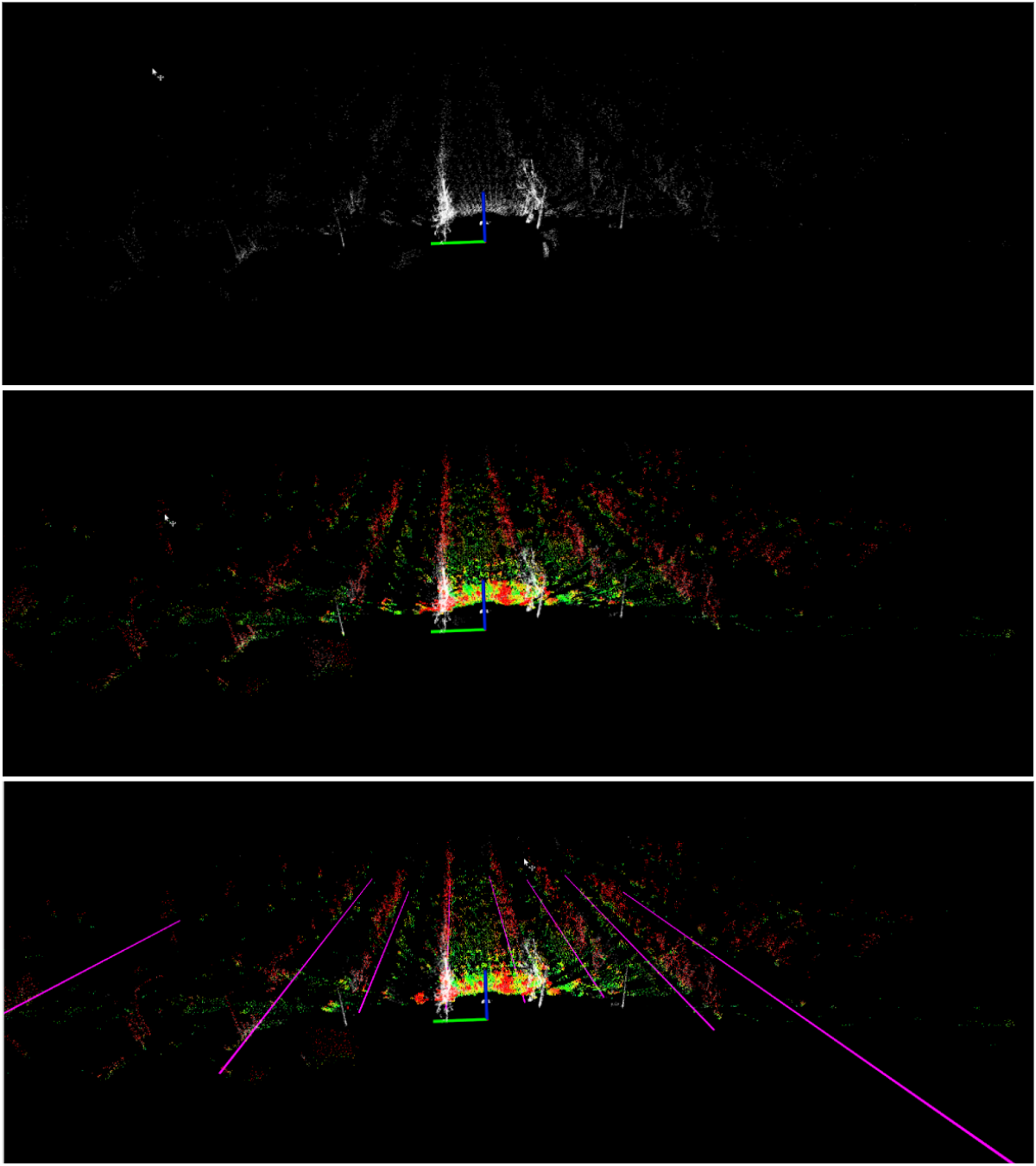}
  \caption{LiDAR-based line detection pipeline. Top: raw point cloud. Middle: filtered
  points used for detection. Bottom: extracted row boundary segments.}
  \label{fig:line_detection_pipeline}
\end{figure}

The line detection module extracts candidate vineyard row boundaries from LiDAR point
clouds by modeling rows as linear structures in the horizontal plane
(Fig.~\ref{fig:line_detection_pipeline}). Ground returns are first suppressed using an
intensity threshold, and an optional rectangular spatial filter clips the point cloud to
a fixed lateral band in the robot frame to exclude adjacent rows. Remaining points are
projected to 2D and passed to an iterative RANSAC fitting procedure.

At each RANSAC iteration, a candidate line is defined by two randomly sampled points.
Inliers are identified by their perpendicular distance to the candidate line,
\begin{equation}
    d_i = \left| (\mathbf{x}_i - \mathbf{p}_0) \cdot \hat{\mathbf{n}} \right|,
\end{equation}
where $\mathbf{p}_0$ is a point on the line and $\hat{\mathbf{n}}$ is the unit normal.
The best model over all iterations is accepted if its inlier count exceeds $N_{\min}$;
accepted inliers are removed and the process repeats until up to $N_{\text{lines}}$
segments are extracted. Each accepted model is converted to a finite segment by
projecting inliers onto the line direction and taking the minimum and maximum extents.

\subsection{Row Exploration \& Navigation}

The row exploration module implements a four-state finite-state machine —
\textit{Initialising}, \textit{Selecting Row}, \textit{Navigating to Row}, and
\textit{In Row} — with transitions governed by geometric predicates evaluated against
the robot pose and current line segments.

\textbf{Row selection.} All pairs of detected segments are tested for approximate
parallelism and minimum separation. Parallelism is accepted when
\begin{equation}
    \left| \hat{\mathbf{d}}_1 \cdot \hat{\mathbf{d}}_2 \right| > 1 - \tau_{\parallel},
\end{equation}
where $\hat{\mathbf{d}}_1$, $\hat{\mathbf{d}}_2$ are the unit directions of the two
segments and $\tau_{\parallel}$ is a cosine-based tolerance. Among valid candidates
within $d_{\max}$ of the robot, the selected row minimises the mean endpoint distance,
\begin{equation}
    J(\ell_i,\ell_j) = \frac{1}{2}\left(\delta(\mathbf{r},\ell_i) +
    \delta(\mathbf{r},\ell_j)\right),
    \label{eq:row_cost}
\end{equation}
where $\delta(\mathbf{r},\ell_k)$ is the distance from the robot position $\mathbf{r}$
to the nearest endpoint of segment $\ell_k$. This favours locally observable rows and
reduces sensitivity to distant, partially observed structure. 

During initial row acquisition, $\mathbf{r}$ in Eq.~\ref{eq:row_cost} is taken as the
robot's SLAM pose. After a completed traversal, however, evaluating $J$ at the
SLAM pose would re-select the just-traversed row, since it remains the geometrically
nearest candidate. To enable sequential lane-to-lane exploration, $J$ is instead
evaluated at a padded query position
\begin{equation}
    \mathbf{r}' = \mathbf{r} + w_{\text{row}}\,\hat{\mathbf{n}},
    \label{eq:padded_query}
\end{equation}
where $\hat{\mathbf{n}}$ is the unit normal to the just-traversed row direction (with
sign alternating across successive traversals) and $w_{\text{row}}$ is the row width
estimated from the previous traversal. Since $\mathbf{r}'$ lies geometrically inside
the adjacent lane, $J$ is minimised by the boundary pair enclosing that lane. The
padded position is used only as a query for row selection; once a target row is
selected, in-row navigation generates waypoints from the robot's actual pose.

% Reviewer has a question about the Equ.3 when the robot finishes an in-row navigation -- as according to Equ.3 the robot shall "re-select" the rows it just traversed as they are the closest pair. However we actually mentioned here that "a heuristic biases selection towards the neighboring lane after each completed traversal". I think we can expand this point to make it a bit clearer to the reader.   
% Update: I am actively checking this detail with Spencer. Once I make this clear I will finish this point and email everyone for final check. Pending...

\textbf{In-row traversal.} Once inside a row, the lane centerline is estimated as the
midpoint between the closest points on the two boundary segments, and a waypoint is
placed ahead along the row tangent,
\begin{equation}
    \mathbf{w} = \mathbf{c} + \Delta\hat{\mathbf{t}},
\end{equation}
where $\mathbf{c}$ is the estimated center, $\hat{\mathbf{t}}$ is the signed unit
tangent, and $\Delta$ is the lookahead distance. The row estimate is refreshed every
$T_{\text{update}}$ seconds by searching for a segment pair consistent with the current
heading, width, and proximity; if no match is found the previous estimate is retained.
The robot exits the row when its position no longer projects within the boundary extents,
triggering a return to \textit{Selecting Row}.

Table~\ref{tab:params} summarizes the key parameters used across all field trials.

\begin{table}[t]
\centering
\caption{Key system parameters used in field experiments.}
\label{tab:params}
\begin{tabular}{l l l}
\hline
\textbf{Parameter} & \textbf{Description} & \textbf{Value} \\
\hline
$I_{\min}$              & Ground intensity threshold        & 0.15  \\
$\epsilon_{\text{ransac}}$ & RANSAC inlier threshold (m)    & 0.2   \\
$N_{\min}$              & Minimum inliers per line          & 250   \\
$N_{\text{lines}}$      & Maximum lines detected            & 8     \\
\hline
$w_{\min}$              & Minimum row width (m)             & 1.0   \\
$w_{\text{row}}$        & estimated row spacing (m)           & 2.5   \\
$\tau_{\parallel}$      & Parallelism tolerance             & 0.2   \\
$d_{\max}$              & Maximum selection distance (m)    & 10.0  \\
$\epsilon_{\text{exit}}$& Row exit threshold (m)            & 1.0   \\
\hline
$\Delta$                & Waypoint lookahead distance (m)   & 2.0   \\
$T_{\text{update}}$     & Waypoint update interval (s)      & 10.0  \\
\hline
\end{tabular}
\end{table}

\section{EXPERIMENTS \& RESULTS}

\subsection{Deployment Setup}

The system was evaluated during a single-day field deployment at Denbies Vineyard (UK) on November~28, 2025 (Fig.~\ref{fig:platform_in_vineyard}). The test environment
consisted of vineyard lanes with approximately $2.5\,\mathrm{m}$ row spacing and row lengths of at least $185\,\mathrm{m}$. Terrain conditions were representative of late-season operation: the ground was predominantly grass and mud with a slope of ${\sim}5^\circ$. Vegetation was sparse, with no leaves or grapes present, reducing the continuity of row boundary structure compared to peak-season canopy conditions.
Approximately two hours of on-site testing time were available, necessitating an iterative trial structure with system parameters updated between runs. All trials were executed in fully autonomous mode, with an operator present solely as a safety
intervention mechanism. A video of the experimental results is available at: \url{youtu.be/1om1HFGF4ps}

\subsection{Overall Autonomous Traversal Performance}

\begin{figure}[t]
  \centering
  \includegraphics[width=\linewidth]{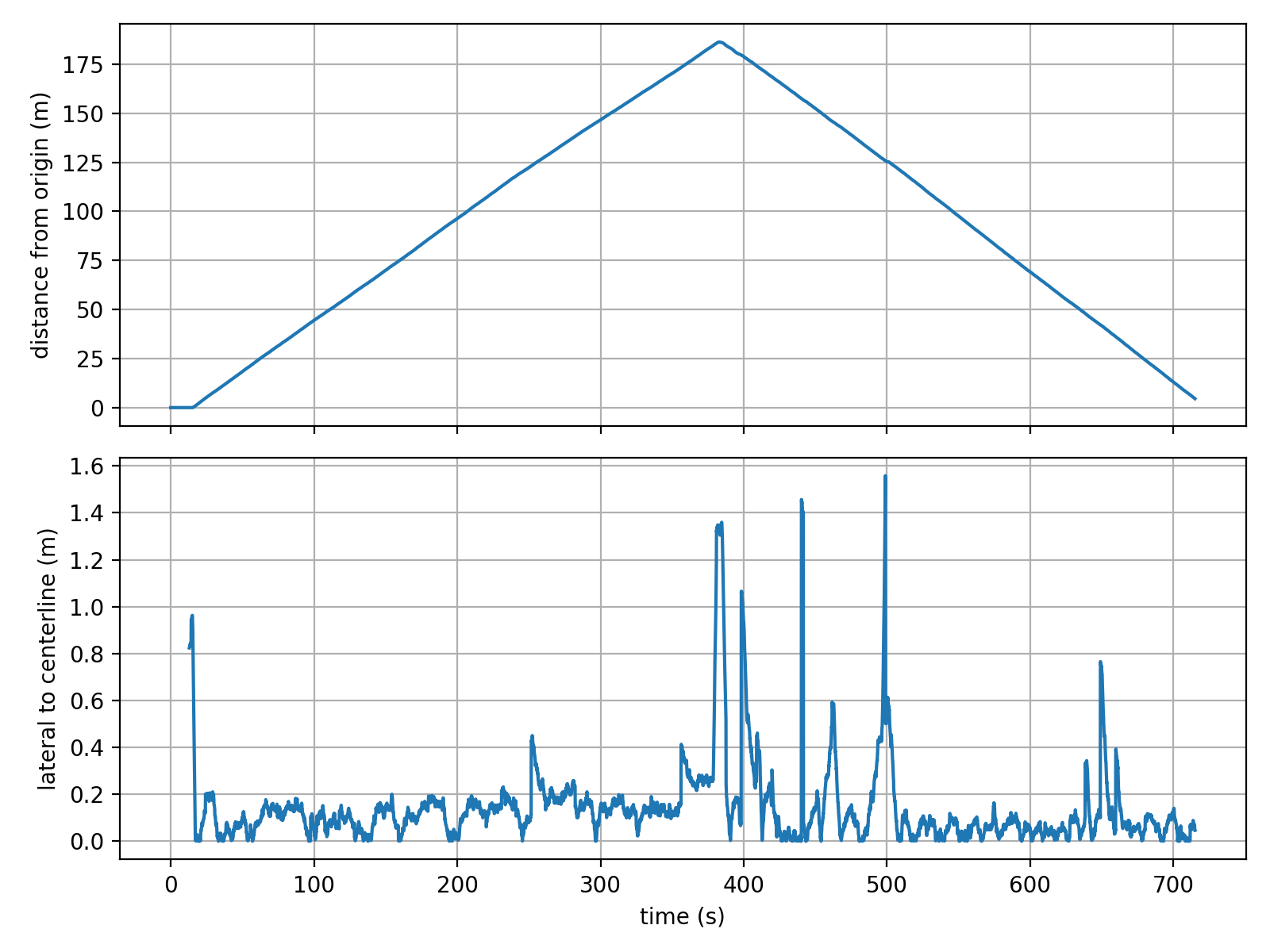}
  \caption{Robot motion during the successful autonomous traversal. Top: displacement
  from origin. Bottom: lateral offset from the estimated row centerline (mean
  $0.14\,\mathrm{m}$). Spikes correspond to obstacle avoidance maneuvers and brief
  row geometry tracking errors.}
  \label{fig:distance_and_lateral_offset}
\end{figure}

The system achieved one fully autonomous traversal, covering $370\,\mathrm{m}$ in
approximately $11\,\mathrm{min}$, completing an up-and-down pass within a single
vineyard row without operator intervention. Figure~\ref{fig:distance_and_lateral_offset}
shows the displacement profile and lateral offset from the estimated row centerline.
The mean lateral offset across the full traversal was $0.14\,\mathrm{m}$, compared with estimated row space (approximate 2.5 m) indicating
consistent in-row tracking. We emphasize that this offset is measured against the \emph{instantaneous} centerline estimate. Therefore, transient spikes either reflects obstacle avoidance maneuvers around patches of high grass, or brief tracking errors from instability of row update (see \ref{sec:discuss-failure-case}). The robot itself
remained physically within its lane throughout the autonomous traversal without operator
intervention.

\subsection{Optimization of Line Detection Stability}
Achieving the continuous traversal required on-site optimization of the perception
pipeline to address environmental noise and structural ambiguity in vineyard rows.

\textbf{RANSAC Parameter Tuning.} Initial trials revealed that limiting the maximum
number of detected line segments frequently led to misclassification of the active row,
as geometrically well-supported but distant rows outcompeted the sparsely observed
boundaries of the current lane. Increasing $N_{\text{lines}}$ preserved weaker but
locally relevant boundaries, significantly reducing active-row dropout.

\textbf{Spatial Filtering of LiDAR Observations.} Residual interference from adjacent
rows was suppressed by applying a rectangular spatial filter in the robot's local frame.
Unlike a circular region of interest, this asymmetric design selectively removes lateral
neighbours without truncating forward visibility, and proved to be a key enabler of the
successful $370\,\mathrm{m}$ traversal.

\subsection{Failure Cases and Limitations}
\label{sec:discuss-failure-case}
\begin{figure}[t]
  \centering
  \includegraphics[width=\linewidth]{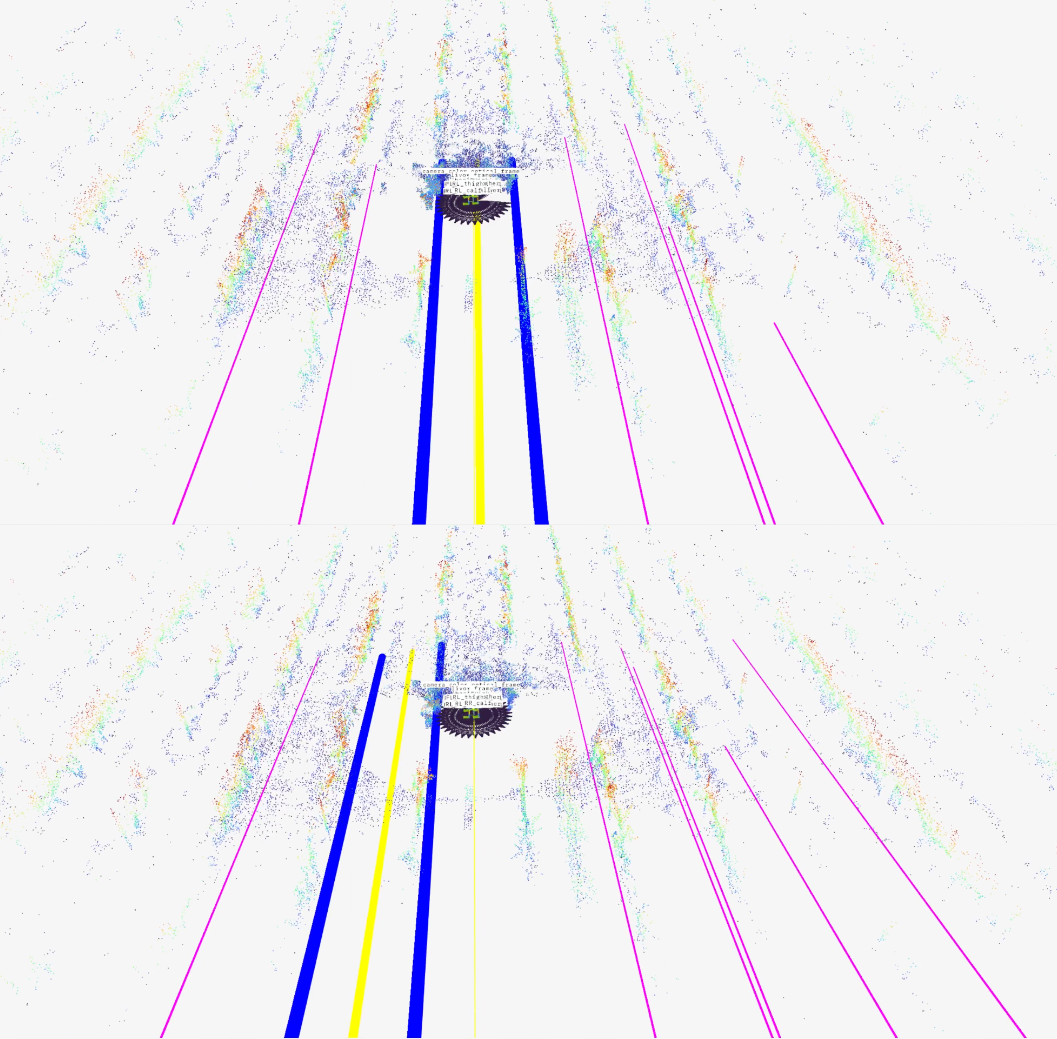}
  \caption{Row update instability. Blue lines mark the currently selected
  row boundaries, the yellow line is the estimated row centerline, purple
  lines are other detected line segments, and the rainbow-coloured points
  are the raw LiDAR point cloud. Top: correct detection of both row
  boundaries. Bottom: one boundary is lost due to a vegetation gap,
  causing the system to update its row estimate to the adjacent lane}
  \label{fig:row_update_failure}
\end{figure}

\textbf{Row Update Instability in Sparse Vegetation.} The most significant challenge
occurred in regions with sparse or discontinuous vegetation. When a vegetation gap coincided
with a row update event, the system occasionally switched its internal representation
to an adjacent row exhibiting a stronger geometric fit. This behaviour arises from a
geometric imbalance: the active sparce row might accumulate fewer RANSAC inliers than neighbouring rows, so the scoring criterion favours an adjacent lane despite its greater lateral distance. The fixed update interval $T_{\text{update}}$ exacerbates this by committing
to a new row estimate at regular intervals regardless of detection confidence.
Figure~\ref{fig:row_update_failure} illustrates this failure. Although the system sometimes
self-corrected when vegetation reappeared and the active row regained sufficient inlier
density, manual intervention was required during preliminary trials to reset the robot's
heading.

\textbf{Row Transition Failure.} Autonomous lane-to-lane transitions were attempted
with the spatial filter disabled, as the filter constrained the lateral visibility
required to observe the target row during turning. Without the filter, line detection
instability prevented reliable lock-on to the target row, exposing a circular
dependency inherent to the current design: disabling the filter is necessary to perceive
the target row, yet doing so destabilises the current row estimate. Furthermore, the
system maintains only a single row hypothesis at any given time and therefore cannot
simultaneously track the current lane and reason about the target lane during a maneuver.
Consequently, no reliable autonomous row transitions were achieved during the trials.

\subsection{Discussion}

The primary limitation identified in field trials is the memoryless nature of the
perception pipeline. Because the system processes each LiDAR frame independently, it
lacks temporal continuity to bridge vegetation gaps, leaving the system vulnerable to
misclassification at update events. The rectangular spatial filter mitigates this for
straight traversal by artificially constraining the field of view, but introduces a
significant trade-off: the same rigid constraint that reduces perceptual ambiguity
during in-row navigation prevents the robot from perceiving the lateral context required
for row transitions, directly resulting in the failure of autonomous turning maneuvers.
Both failure modes are therefore manifestations of the same underlying issue, namely the
absence of any confidence measure on detections and any memory between frames. Resolving
this requires a perception strategy sensitive to navigation context, maintaining tighter
spatial constraints during traversal and relaxing them adaptively during turning
maneuvers.

\section{CONCLUSION}

We presented a LiDAR-based autonomous exploration system for vineyard environments,
combining iterative RANSAC-based line detection with a finite-state row navigation
strategy, and evaluated its performance during a real-world field deployment on a
Unitree Go2-W. Field experiments at a commercial vineyard demonstrated a fully
autonomous $370\,\mathrm{m}$ up-and-down row traversal without operator intervention
under challenging late-season conditions. While the system proved effective for
long-distance row following, intermittent line detection instability under sparse
vegetation was identified as the primary failure mode, affecting both row updates and
lane-to-lane transitions. A rectangular spatial filtering strategy improved in-row
stability but constrained visibility of adjacent rows, illustrating an inherent
trade-off between robustness and exploratory capability. These results highlight key
challenges for reliable autonomous vineyard navigation and motivate future work on
temporally-aware, context-sensitive perception strategies.  Future evaluations will 
incorporate repeated trials and quantitative characterisation of failure rates 
(e.g., row update instability events per kilometre) to support reliability assessment.

\section*{ACKNOWLEDGMENT}
This work has been partially supported by project MIS 5154714 of the National Recovery and Resilience Plan Greece 2.0 funded
by the European Union under the NextGenerationEU Program.

\addtolength{\textheight}{-12cm}   % This command serves to balance the column lengths

\bibliographystyle{IEEEtran}
\bibliography{ref}

\end{document}